\documentclass[10pt,twocolumn,letterpaper]{article}

\usepackage[pagenumbers]{cvpr} 

\usepackage{graphicx}
\usepackage{amsmath}
\usepackage{amssymb}
\usepackage{booktabs}
\usepackage{times}
\usepackage{epsfig}
\usepackage{multirow}
\usepackage{bm}
\usepackage{enumitem}
\usepackage{arydshln}
\usepackage{adjustbox}
\usepackage{cuted}
\usepackage[colorlinks,linkcolor=blue]{hyperref}
\usepackage[accsupp]{axessibility}
\usepackage[capitalize]{cleveref}
\crefname{section}{Sec.}{Secs.}
\Crefname{section}{Section}{Sections}
\Crefname{table}{Table}{Tables}
\crefname{table}{Tab.}{Tabs.}

\def\cvprPaperID{3630} 
\def\confName{CVPR}
\def\confYear{2023}

\newcommand{\nothing}[1]{}

\begin{document}

\title{Hand Avatar: Free-Pose Hand Animation and Rendering from Monocular Video
}

\author{Xingyu Chen \hspace{0.15in}
Baoyuan Wang \hspace{0.15in}
Heung-Yeung Shum \\
Xiaobing.AI \\
}

\maketitle

\begin{strip}
    \centering
    \vspace{-10pt}
    \includegraphics[width=\linewidth]{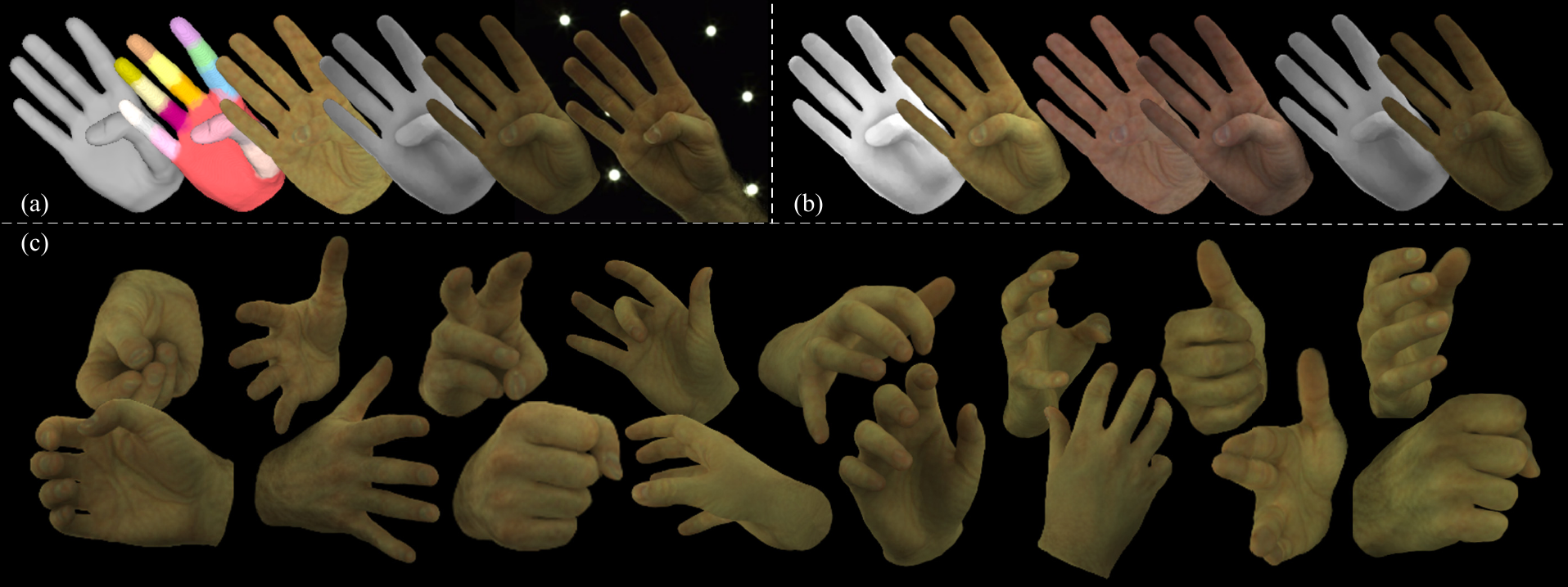}\\
    \captionsetup{type=figure,font=small}
    \caption{Demonstration of HandAvatar. (a) Personalized hand rendering. From left to right: hand mesh, compositional occupancy, albedo, illumination, shaded appearance, and ground truth; (b) three groups of texture editing in terms of lighting, albedo, and shadow (by altering the self-occlusion effect of thumb); (c) free-pose hand animation and rendering.
    }
    \label{fig:fig1}
\end{strip}

\begin{abstract}
We present HandAvatar, a novel representation for hand animation and rendering, which can generate smoothly compositional geometry and self-occlusion-aware texture. Specifically, we first develop a MANO-HD model as a high-resolution mesh topology to fit personalized hand shapes. Sequentially, we decompose hand geometry into per-bone rigid parts, and then re-compose paired geometry encodings to derive an across-part consistent occupancy field. As for texture modeling, we propose a self-occlusion-aware shading field (SelF). In SelF, drivable anchors are paved on the MANO-HD surface to record albedo information under a wide variety of hand poses. Moreover, directed soft occupancy is designed to describe the ray-to-surface relation, which is leveraged to generate an illumination field for the disentanglement of pose-independent albedo and pose-dependent illumination. Trained from monocular video data, our HandAvatar can perform free-pose hand animation and rendering while at the same time achieving superior appearance fidelity. We also demonstrate that HandAvatar provides a route for hand appearance editing. 
Project website: \url{https://seanchenxy.github.io/HandAvatarWeb}.
\end{abstract}

\vspace{-0.5cm}
\section{Introduction}
Human avatars \cite{bib:NeuMan,bib:NHA,bib:IMAvatar,bib:AVA,bib:MVP,bib:SelfRecon} have been vigorously studied for years. However, there has been limited research that particularly focuses on hand avatars \cite{bib:LISA}. Due to the nature of distinctive properties (\textit{e.g.}, serious self-occlusion and contact) between the hand and the rest of the human parts (\textit{i.e.}, face, head, and body), it is essential to investigate a specialized representation tailored for modeling both the hand geometry and texture.

Traditional pipeline tends to adopt texture maps and colored mesh for hand appearance modeling \cite{bib:DART,bib:HTML,bib:NIMBLE,bib:S2HAND,bib:CycleHand}, but developing an elaborate personalized hand mesh and texture map usually requires expensive scan data \cite{bib:ten24} and artistic knowledge. Recently, the neural rendering technique has gained raising attention, where neural radiance field (NeRF) \cite{bib:NeRF} has been adapted to represent humans by predicting geometry and texture properties for an arbitrary 3D point query \cite{bib:SLRF,bib:NeuralActor,bib:NeuMan,bib:STNeRF,bib:HumanNeRF,bib:HumanNeRFSP,bib:AnimNeRF,bib:NeuralBody,bib:SANeRF,bib:GRAM,bib:DoubleField,bib:LISA,bib:ANeRF,bib:HNeRF,bib:NRNeRF,bib:Nerfies} . Compared to the conventional mesh-texture pipeline, NeRF is cheap in training data collection and superior in rendering fidelity. Despite the huge success of human body and face modeling, neural rendering-based hand representation \cite{bib:LISA} remains much less explored. The hand is highly articulated such that the complex hand motion brings difficulties for neural rendering. Firstly, the deformation of hand geometry is hard to model. When coping with large and complex hand deformations (\textit{e.g.}, self-contact), previous skinning-based methods can hardly find accurate skinning weights for an arbitrary query \cite{bib:InvLBS,bib:HumanNeRF,bib:AnimNeRF,bib:NARF,bib:loopreg,bib:SelfRecon,bib:snarf,bib:IMAvatar,bib:LEAP,bib:SCANimate}, while part-aware methods usually suffer from across-part inconsistency issue \cite{bib:NASA,bib:HALO,bib:COAP,bib:PTF}. Secondly, hand texture is hard to model because of the highly articulated structure. For example, articulated hand motion induces serious self-occlusion so that different hand poses lead to noticeable variations in illumination and shadow patterns. Illumination is important for realistic rendering, but we are not aware of any prior work in estimating illumination caused by articulated self-occlusion. 

Motivated by the above challenges, we propose HandAvatar for animatable realistic hand rendering. Considering different difficulties in geometry and texture modeling, we follow the idea of inverse graphics \cite{bib:HyFace} to disentangle hand geometry, albedo, and illumination. At first, we employ explicit mesh to depict hand shapes. However, the popular hand mesh model, \textit{i.e.}, MANO \cite{bib:MANO}, only provides a coarse mesh with 778 vertices, whose shape fitting capacity is limited. Therefore, we design a super-resolution version of MANO with 12,337 vertices and 24,608 faces, namely MANO-HD, which can fit personalized hand shapes with per-vertex displacements. Additionally, massive existing MANO-annotated data can be seamlessly represented by MANO-HD. For introducing mesh-based hand shape to the volume rendering pipeline \cite{bib:NeRF}, we propose a local-pair occupancy field (PairOF), where every two part-level geometry encodings are reassembled according to physical connections to yield an across-part consistent field. As for hand texture, we propose a self-occlusion-aware shading field (SelF). SelF is comprised of an albedo field and an illumination field. The albedo field resorts to anchors that are uniformly paved on MANO-HD surfaces, each of which holds positional and albedo encodings to model a small hand region. The illumination field is to cope with articulated self-occlusion, where directed soft occupancy is designed to estimate illumination and shadow patterns. 

MANO-HD and PairOF are pre-trained with MANO parameter annotations, then they cooperate with SelF in end-to-end training on monocular video data. Finally, with hand pose as the input, our HandAvatar can perform hand animation and rendering. We evaluate our approach on the InterHand2.6M dataset \cite{bib:InterHand} and achieve high-fidelity geometry and texture for free-pose hand animation. We also demonstrate that it is convenient to edit hand appearance in HandAvatar as shown in Fig.~\ref{fig:fig1}. Therefore, our main contributions are summarized as follows:
\begin{itemize}[leftmargin=*,topsep=3pt]
    \setlength\itemsep{0em}
    \item We propose a HandAvatar framework, the first method for neural hand rendering with self-occluded illumination.
    \item We develop MANO-HD and a local-pair occupancy field that fit hand geometry with personalized shape details.
    \item We propose a self-occlusion-aware shading field that can render hand texture with
    faithful shadow patterns.
    \item Our framework is end-to-end developed for free-pose realistic hand avatars. Extensive evaluations indicate our method outperforms prior arts by a large margin.
\end{itemize}

\section{Related Work}
\paragraph{Articulated Human Geometry.}
Parametric human models \cite{bib:SMPL,bib:SMPLX,bib:MANO,bib:FLAME} have developed for years, where mesh can be inferred given pose and shape parameters. Specifically, as a common-used hand model, MANO \cite{bib:MANO} can produce a hand mesh with 778 vertices and 1,538 faces. This mesh template is too coarse so its representation capacity is largely limited. Gyeongsik \textit{et al.} \cite{bib:DeepHandMesh} proposed DeepHandMesh to generate dense and high-fidelity hand mesh, but brought restricted generalization as multi-view depth data was required for training. In contrast, our MANO-HD is a general high-resolution hand mesh model so that all existing MANO-annotated data can be seamlessly represented using MANO-HD. Meanwhile, MANO-HD can fit personalized hand shapes with monocular RGB video data.

Mesh suffers drawbacks of discontinuity and unalterable topology structure. To remedy this issue, recent research tends to explore implicit human geometry \cite{bib:GraspField,bib:PIFu,bib:LEAP,bib:imGHUM,bib:imFace,bib:i3DMM}, which has the advantages of flexibility and continuity. For example, GraspField \cite{bib:GraspField} leveraged the signed distance field (SDF) to describe hand-object contact. However, implicit geometry is poor in free-pose animation when compared to explicit mesh, so the articulated driving of implicit human geometry is widely studied. As reported in \cite{bib:HumanNeRF,bib:AnimNeRF,bib:NARF,bib:loopreg}, a posed-space query can be transformed back to canonical space with linear blend skinning and inverse skinning weights. The inverse skinning paradigm fails to deal with self-contact, where a query can match multiple canonical-space points. Then, forward skinning deformation is designed to transform canonical-space points to posed space with an iterative root finding method \cite{bib:SelfRecon,bib:IMAvatar,bib:LEAP,bib:snarf,bib:SCANimate}, but the iterative optimization algorithm could hurt end-to-end network training. By and large, per-bone rigid transformations can compose a large motion space with the difficulty of optimizing accurate skinning weights for an arbitrary 3D point query. With the aid of parametric models \cite{bib:SMPL,bib:MANO}, another idea of deformation between posed and canonical spaces is to leverage the surface motion \cite{bib:Te,bib:NeuralActor,bib:SANeRF}. For a query, a mesh-surface point is found according to Euclidean distance as the reference, then the query deformation is set the same as that of the reference point. Yuan \textit{et al.} \cite{bib:NeRFEdit} and Garbin \textit{et al.} \cite{bib:voltemorph} pointed out that the reference from triangular mesh is not accurate enough and proposed to use tetrahedral mesh. However, the deformation of tetrahedral mesh is hard to cooperate with popular human priors \cite{bib:SMPL,bib:MANO,bib:FLAME} and could be potentially slow \cite{bib:ARAP}. Without requiring motion approximation, part-aware methods \cite{bib:NASA,bib:HALO,bib:COAP,bib:PTF} are developed by fusing part-wise geometries. NASA \cite{bib:NASA} divided the body into per-bone parts, then a query was deformed into each part space with unambiguous rigid transformation for decoding of part-level occupancy. NASA can describe complex deformations owing to accurate query motion, but information between body parts is ignored. To relieve this issue, COAP \cite{bib:COAP} encoded connected parts with PointNet \cite{bib:PointNet}, but it still incurred non-smooth part connections. Instead of part-wise modeling, we propose a part-pair-wise decoder to generate across-part consistent geometry.

\vspace{-0.3cm}
\paragraph{Human Texture.}
Previously, the primary focus on hand texture is the texture map and colored mesh \cite{bib:DART,bib:HTML,bib:NIMBLE,bib:S2HAND,bib:CycleHand}. Although many high-quality texture maps are explored, the design of personalized texture maps usually requires expensive scan data \cite{bib:ten24} and artistic knowledge \cite{bib:DART}. In contrast, LISA \cite{bib:LISA} employed radiance field \cite{bib:VolSDF} to learn hand appearance from multi-view images and introduced color parameters for texture generalization. Different from LISA, we design a monocular method for the convenience of data collection. Moreover, we pay attention to detailed personalized textures including albedo and illumination. Because of the aforementioned difficulty in implicit deformation, the learned texture on the human surface could be blurred \cite{bib:HumanNeRF,bib:SelfRecon}. To enhance surface texture representation, local representations are developed with explicit mesh as the guidance. NeuralBody \cite{bib:NeuralBody} attached latent codes to mesh vertices, which can diffuse into space with sparse convolution \cite{bib:SparseConv}. NeuMesh \cite{bib:NeuMesh} also put color features on mesh vertices, and achieved an editable radiance field. Furthermore, mesh-guided local volume \cite{bib:MVP} and local radiance field \cite{bib:SLRF} were designed. We follow the local modeling paradigm and uniformly place anchors on MANO-HD surface using barycentric sampling to trace local information. 

\vspace{-0.3cm}
\paragraph{Human Inverse Rendering.} 
Most methods model human appearance with entangled geometry, albedo, and illumination \cite{bib:HumanNeRF}. Meanwhile, there has been a surge of interest in human inverse rendering, the purpose of which is to extract intrinsic components (\textit{i.e.}, geometry, material, and illumination) from RGB data \cite{bib:Tewari21,bib:fml,bib:gan2x,bib:mofa,bib:HyFace}. For example, GAN2X
\cite{bib:gan2x} designed an unsupervised framework to model albedo and specular properties of non-Lambertian material, then rendered face with Phong shading \cite{bib:phong}. With a similar purpose, HyFRIS-Net \cite{bib:HyFace} disentangled albedo and illumination with an inverse 3DMM model to achieve a considerably improved quality of face rendering. S2HAND \cite{bib:S2HAND} simultaneously estimated camera pose, colored mesh, and lighting condition to form a photometric loss for hand pose estimation, but its rendering quality was coarse without a detailed appearance. Although the inverse rendering technique on the human face has been becoming a well-studied issue, the knowledge cannot be trivially transferred to hand tasks. Different from the face, the hand is characterized by articulated self-occlusion. Illumination and shadow caused by self-occlusion have not yet been discussed in prior works, and thus we fill this gap for hand inverse graphics.

\vspace{-0.3cm}
\paragraph{Illumination in Radiance Field.}
The existing literature on NeRF-based illumination technique \cite{bib:NeRD,bib:NeRV,bib:NeLF,bib:NeRFactor} is to estimate source light condition or surface reflection property (\textit{i.e.}, bidirectional reflectance distribution function, BRDF). For example, NeRV \cite{bib:NeRV} took as input a set of images under known lighting to predict BRDF, and achieved novel-view rendering with arbitrary lighting conditions. NeLF \cite{bib:NeLF} designed a lighting estimation module, and then performed face relighting. NeRF-OSR \cite{bib:NeRFOSR} collected multi-view outdoor images to predict albedo and shadow maps. In contrast to the prior art, we dedicate modeling illumination under the condition of articulated self-occlusion.

\section{Method}

\begin{figure*}[t]
\begin{center}
\includegraphics[width=\linewidth]{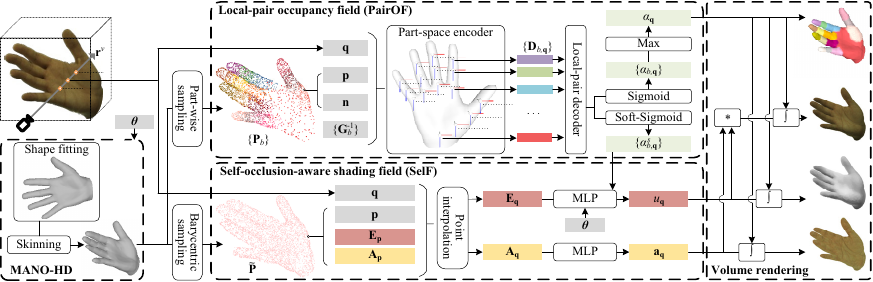}
\caption{HandAvatar overview. Given hand pose, MANO-HD produces personalized mesh, while PariOF yields accordingly occupancy field. SelF estimates albedo and illumination fields under self-occlusion. Then, hand appearance is synthesized by volume rendering.
}
\label{fig:arch}
\end{center}
\vspace{-0.3cm}
\end{figure*}

Fig.~\ref{fig:arch} illustrates the overview pipeline of our HandAvatar system, including MANO-HD (Sec.~\ref{sec:manohd}), PairOF (Sec.~\ref{sec:pairof}), and SelF (Sec.~\ref{sec:self}). Table~\ref{tab:symbol} also provides the list of symbol notations and their definitions used in this paper.

\subsection{MANO-HD}
\label{sec:manohd}

\begin{figure}[t]
\begin{center}
\includegraphics[width=\linewidth]{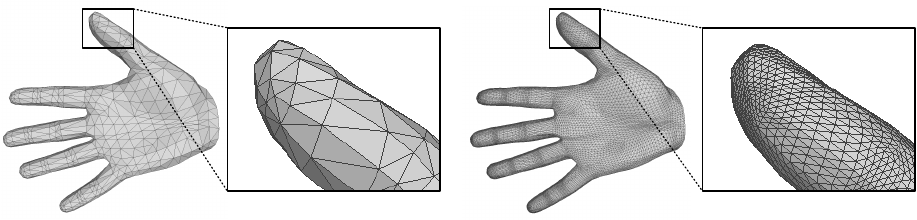}
\caption{Templates of MANO (left) and MANO-HD (right).}
\label{fig:temp}
\end{center}
\vspace{-0.5cm}
\end{figure}

\paragraph{Mesh Subdivision.}
MANO \cite{bib:MANO} deforms hand mesh with shape parameter $\boldsymbol{\beta}$ and pose parameter $\boldsymbol{\theta}\in\mathbb R^{B\times 3}$ ($B=16$ indicates the number of per-bone parts). For lifting mesh resolution, we uniformly subdivide MANO template mesh by adding new vertices on edge middle points \cite{bib:NHA}. This operation increases the vertex number to 12,337 and the face amount to 24,608 (see Fig.~\ref{fig:temp}). 
Then, the skinning weights of added vertices are given with the average of their seminal vertices. 
To eliminate artifacts during skinning, we optimize upsampled skinning weights for better dynamic performance. Please see the \textit{suppl. material} for details.

\begin{table}[t]
\small
\setlength\tabcolsep{3pt}
\centering
\begin{tabular}{c c | c c}
\hline
$\boldsymbol\theta$ & hand pose parameter & $\boldsymbol\beta$ & hand shape parameter \\
$\mathbf P$ & point cloud & $\mathbf r$ & ray direction \\
$b$ & index of per-bone parts  & $\mathbf n$ & sampled point normal \\
$\mathbf q$ & query point position & $\mathbf p$ & sampled point position \\
$\mathbf G_b$ & bone transformation matrix & $\mathbf D_b$ & part geometry encoding \\
$\mathbf E$ & positional encoding & $\mathbf A$ & albedo encoding \\
$u$ & illumination value & $\mathbf a$ & albedo value in RGB \\
$\mathbf \alpha$ & occupancy value & $\alpha^s$ & soft occupancy value \\
\hline
\end{tabular}
\caption{Symbolical notations.}
\label{tab:symbol}
\end{table}


\vspace{-0.2cm}
\paragraph{Shape Fitting.}
Although MANO-HD has a high-resolution template, its shape is still bounded by $\boldsymbol{\beta}$. Hence, when modeling personalized hand mesh, we get rid of $\boldsymbol{\beta}$ and use a multi-layer perceptron (MLP) to derive a refined shape $\tilde{\mathbf V} =\bar{\mathbf V}+\mathcal M_{shape}([\mathcal P(\bar{\mathbf V}),\boldsymbol\theta])$,
where $\bar{\mathbf V},\mathcal P(\cdot),[\cdot]$ denote the MANO-HD template vertices, positional encoding, and concatenation. 
The MLP can be trained with monocular video data and IoU loss $\mathcal L_{shape} = 1-\text{IoU}(\mathcal D(\mathcal S(\tilde{\mathbf V})), \mathbf S^*)$, where $\mathcal D,\mathcal S,\mathbf S^*$ are silhouette rendering \cite{bib:SoftRas}, linear blend skinning, and ground-truth silhouette.

\subsection{Local-Pair Occupancy Field}
\label{sec:pairof}
Given the query point $\mathbf q\in\mathbb R^3$, PairOF predicts the occupancy value $\alpha_\mathbf q$ to describe whether it locates in ($\mathbf {\alpha_q}>0.5$) or out of ($\mathbf {\alpha_q}<0.5$) the surface. Hence, the hand surface can be formulated as $\{\mathbf q|\mathbf{\alpha_q}=0.5\}$. 

\vspace{-0.3cm}
\paragraph{Part-Space Encoder.}
Following NASA \cite{bib:NASA}, we divide hand mesh into per-bone part meshes and uniformly sample $N^p$ points on part mesh faces to obtain point clouds $\mathbf P_b=\{\mathbf p\in\mathbb R^3\}$ and normals $\mathbf N_b=\{\mathbf n\in\mathbb R^3\}$. Points, normals, and query are transferred back to part canonical spaces with $\hat{\mathbf P}_b=\mathbf G^{-1}_b\mathbf P_b, \hat{\mathbf N}_b=\mathbf R^{-1}_b\mathbf N_b, \hat{\mathbf q}_b=\mathbf G^{-1}_b\mathbf q$, where $\mathbf G$ is bone transformation matrix and $\mathbf R$ is the rotation component of $\mathbf G$. Following COAP \cite{bib:COAP}, $\hat{\mathbf P}_b,\hat{\mathbf N}_b$ are fed to a PointNet $\mathcal Q_{part}$ to extract latent geometry features. The part geometry encoding is ultimately formulated by concatenating PointNet representation and canonicalized query, \textit{i.e.}, $\mathbf D_{b,\mathbf q}=[\mathcal Q_{part}(\hat{[\mathbf P}_b, \hat{\mathbf N}_b]),\hat{\mathbf q}_b]$.

\vspace{-0.3cm}
\paragraph{Local-Pair Decoder.}
To explore inter-part relations, some previous works fuse per-bone features according to the kinematic tree to form structured representations \cite{bib:Georgakis,bib:Aksan,bib:LEAP}. Our intuition is that the local part shape is not related to the kinematic tree, instead, there is a strong relationship between 
parts. Hence, we define local pair as two parts that are physically connected. Then, we propose local-pair decoder $\mathcal Q_{pair}$ based on PointNet to fuse each paired encodings and predict occupancy value:
\begin{equation}
\label{eq:pair}
\begin{array}{l}
\alpha_{b,\mathbf q} = \sigma(\max\{\mathcal Q_{pair}(\{\mathbf D_{b,\mathbf q}, \mathbf D_{b',\mathbf q}\})| b'\in\mathbb P(b)\})
\end{array}
\vspace{-0.1cm}
\end{equation}
where $\mathbb P(b)$ selects locally paired parts that has a physical connection with part $b$; $\sigma$ is the sigmoid function; $\alpha_{b,\mathbf q}$ is part-level occupancy value. Through the part-pair-wise decoding, the part boundaries become blurred. Therefore, the intuition behind the maximum operator in Eq.~(\ref{eq:pair}) is to yield a union of part-level geometries that extend to the connections. Finally, The global occupancy value is given by fusing part-level values, \textit{i.e.}, $\alpha_{\mathbf q} = \max\{\alpha_{b,\mathbf q}\}_{b=1}^B.$

\vspace{-0.3cm}
\paragraph{Pre-Training.}
With free MANO parameter annotations, we pre-train PairOF as a prior model to endow PairOF with prior knowledge of 3D hands. With a hand mesh inferred by MANO-HD, we sample point clouds with $N^t$ points as training data \cite{bib:COAP}.
The objective is to minimize the mean squared error between ground truth $\alpha^*$ and predicted occupancy values, \textit{i.e.}, $\mathcal L_{PairOF}=\frac{1}{N^t}\sum_{\mathbf q} (\alpha_{\mathbf q} - \alpha_{\mathbf q}^*)^2$.

\subsection{Self-Occlusion-Aware Shading Field}
\label{sec:self}

Given a query $\mathbf q$, SelF predicts its albedo and illumination values, which are rendered with the volumetric method.

\vspace{-0.25cm}
\paragraph{Volume Rendering.}
For a ray casting on view direction $\mathbf r^v$, we uniformly sample $N^q$ queries $\{\mathbf q_i\}_{i=1}^{N^q}$, each of which has occupancy $\alpha$, albedo $\mathbf a$, and illumination value $u$. We render neural fields with the volumetric method \cite{bib:NeRF}:
\begin{equation}
\label{eq:render}
\begin{array}{l}
\mathbf X_{\mathbf{r}^v} =\sum_{i=1}^{N^q}(\prod_{j=1}^{i-1}(1-\alpha_{\mathbf q_j})) \alpha_{\mathbf q_i} \mathbf{X}_{\mathbf{q}_{i}}.
\end{array}
\vspace{-0.1cm}
\end{equation}
When $\mathbf X$ equates to $\mathbf a,u,$ or  $u\mathbf a$, we obtain the albedo value, illumination value, or shaded RGB color of a pixel.

\vspace{-0.25cm}
\paragraph{Albedo Field.}
Albedo describes the intrinsic color of the material, which is invariant \textit{w.r.t.} hand pose, illumination, \textit{etc.} Motivated by the invariant property, we fix anchors on the MANO-HD surface, whose relative geodesic locations are independent from varying poses. To this end, we uniformly sample point clouds $\bar{\mathbf P}$ with $N^a$ points on MANO-HD template mesh and represent them with barycentric coordinates. Compared to directly using vertex as the anchor \cite{bib:NeuralBody,bib:NeuMesh,bib:MVP}, our barycentric anchors are more uniform to cover the hand surface. Then, we develop albedo encodings $\mathbf A\in\mathbb R^{N^a\times D^a}$ with random initialization and attach them to anchors. Given hand pose, anchors can be re-sampled based on deformed vertices and fixed barycentric coordinates to form deformed points clouds $\tilde{\mathbf P}$. For a query $\mathbf q$, we find $N^n$ nearest points in $\tilde{\mathbf P}$ and interpolate $\mathbf A$ using inverse Euclidean distances as the weights. Thereby, we obtain the albedo encoding $\mathbf A_\mathbf{q}\in \mathbb R^{D^a}$ and then fed it to an MLP to predict the albedo value $\mathbf a_\mathbf{q}\in \mathbb R^{3}$, \textit{i.e.}, $\mathbf a_{\mathbf q}=\mathcal M_{albedo}(\mathbf A_{\mathbf q})$.

\begin{figure}[t]
\begin{center}
\includegraphics[width=\linewidth]{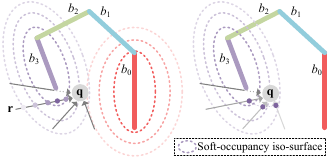}
\caption{Illustration of self-occluded illumination along the kinematic tree of the forefinger. The deepest purple positions contribute to directed soft occupancy value.}
\label{fig:accocc}
\end{center}
\vspace{-0.5cm}
\end{figure}

\vspace{-0.25cm}
\paragraph{Directed Soft Occupancy.}
For self-occluded illumination estimation, we require the near-far relationship for a bone part. That is, the illumination of a query~$\mathbf q$ is affected by self-occlusion when $\mathbf q$ is close to multiple parts. Although occupancy value can describe the relation between $\mathbf q$ and parts, the value is nearly binary so can only depict inside-outside relations. Hence, a soft factor $\tau$ is introduced to the sigmoid function to soften the occupancy value:
\begin{equation}
\label{eq:softocc}
\begin{array}{l}
\sigma^s(x) = \frac{1}{1+e^{-\tau x}}, \quad 0<\tau<1.
\end{array}
\vspace{-0.1cm}
\end{equation}
Soft occupancy $\alpha^s_{b,\mathbf q}$ is derived by replacing $\sigma$ in Eq.~(\ref{eq:pair}) with $\sigma^s$. 
Further, instead of modeling a single query, we design \textit{directed soft occupancy} to reflect the near-far relation between a ray casting and an articulated part. For a ray casting $\mathbf r$ that can reach $\mathbf q$, the directed soft occupancy $\alpha^s_{b,\mathbf q,\mathbf r}$ is defined as the maximal value on $\mathbf r$ before the ray hits $\mathbf{q}$. For discretization, we uniformly sample queries $\{\mathbf q_i\}_{i=1}^{N^q}$ on a ray casting $\mathbf r$, and compute directed soft occupancy as
\begin{equation}
\label{eq:accocc}
\begin{array}{l}
\alpha^s_{b,\mathbf q,\mathbf r} = \max\{\alpha^s_{b,\mathbf q_{i}}|\mathbf q_{i}\leq \mathbf q\},
\end{array}
\vspace{-0.1cm}
\end{equation}
where $\mathbf q_{i}\leq \mathbf q$ selects queries that the ray traverses before reaching $\mathbf q$. For example, $\alpha^s_{b_3,\mathbf q,\mathbf r}$ equals to $\alpha^s_{b_3}$ of the deepest purple query in Fig.~\ref{fig:accocc}.

\vspace{-0.25cm}
\paragraph{Illumination Field.}

It is well known that the illumination effects come with light-source distribution, irradiance, and reflectance. Independent from self-occlusion,  reflectance is the material property, which is not our focus. Affected by self-occlusion, some ambient lighting rays could be occluded such that the irradiance could be changed. Thereby, the problem is formulated as estimating irradiance of an outside query $\mathbf q$ ($\alpha^s_{b,\mathbf q}<0.5$), which indicates the energy amount that can reach $\mathbf q$. To this end, we use the hand pose $\boldsymbol\theta$ and query location as cues. Similar to the albedo encodings, positional encodings $\mathbf E=\mathcal P(\bar{\mathbf P})$ are attached to anchors, and we obtain $\mathbf E_{\mathbf q}$ with interpolation as the surface-calibrated location of $\mathbf{q}$. Nevertheless, self-occlusion is quite intractable for $\boldsymbol\theta$ and $\mathbf {E_q}$, so we leverage directed soft occupancy to enhance the awareness of self-occlusion.

As shown in Fig.~\ref{fig:accocc}, the articulated structure prohibits a portion of energy from arriving $\mathbf q$. Apparently, the situation of energy occlusion around a ray direction is implied in a set of directed soft occupancy $\{\alpha^s_{b,\mathbf q,\mathbf r}\}_{b=1}^B$. That is, if a ray casting is close to multiple parts before hitting $\mathbf q$, the illumination of $\mathbf q$ shall be impacted by self-occlusion. Prohibited by a limited memory budget, we cannot consider spherically distributed ray directions, and thus the number of ray castings is imperative to be reduced. Our institution is that (1) the selected ray should be able to arrive $\mathbf q$ (\textit{i.e.}, $\alpha^s_{b,\mathbf q, \mathbf r}<0.5$) such that can elaborate the near-far relations for all articulated parts; (2) an articulated part can only affect the illumination around it, where the query is close to the part (\textit{i.e.}, $\alpha^s_{b,\mathbf q}\rightarrow 0.5$). Meanwhile, we have $\alpha^s_{b,\mathbf q, \mathbf r}\ge \alpha^s_{b,\mathbf q}$ from Eq.~(\ref{eq:accocc}). Thereby, the variation caused by ray directions is minor, and we use $\{\alpha^s_{b,\mathbf q,\mathbf r^v}\}_{b=1}^B$ as the guidance to estimate irradiance of $\mathbf q$, where $\mathbf r^v$ is the view direction. Without introducing extra ray castings, we significantly reduce computational costs by leveraging the ray casting and queries on the view direction. 

Finally, we use an MLP to predict the illumination value, \textit{i.e.}, $u_{\mathbf q} = \mathcal M_{illum}([\boldsymbol\theta,\mathbf{E_q},[\alpha^s_{b,\mathbf q,\mathbf r^v}]_{b=1}^B])$.

\vspace{-0.25cm}
\paragraph{Optimization.} The training of SelF is based on reconstruction loss functions, including LPIPS \cite{bib:lpips} loss and $l_1$ error between the rendered image $\mathbf C$ and the ground truth $\mathbf C^*$, \textit{i.e.},  $\mathcal L_{SelF}=\mathcal L_{LPIPS}(\mathbf C, \mathbf C^*)+ ||\mathbf C-\mathbf C^*||_1$.

\section{Experiments}
\label{sec:experiments}

\subsection{Implementation Details and Metrics}

\paragraph{Pre-Training of PairOF.}
We adopt all right-hand annotations in InterHand2.6M \cite{bib:InterHand} for pre-training, whose training/test set contains 875,530/565,611 samples. Learnable parameters includes $\mathcal{Q}_{part},\mathcal{Q}_{pair}$. 
We set $N^t=N^p=256$, and the training process is to minimize $\mathcal L_{PairOF}$.

\vspace{-0.3cm}
\paragraph{End-to-End Training.}
With personalized monocular video, we optimize $\mathcal M_{shape},\mathcal Q_{pair},\mathcal M_{albedo}, \mathcal M_{illum}$ and $\mathbf A$ in an end-to-end manner. Video data are selected from InterHand2.6M dataset \cite{bib:InterHand}. 
The objective is to minimize $\mathcal L_{shape}+\mathcal L_{PairOF}+\mathcal L_{SelF}$. Hyperparameters in SelF are set as $N^q=64,N^a=4096,N^n=4,D^a=128,\tau=0.05$. The rendering resolution is $256\times 256$. Please see the \textit{suppl. material} for data selection, pre-processing, and more training details.



\begin{table}[t]
\small
\setlength\tabcolsep{2pt}
\centering
\begin{tabular}{c |c c c c c}
\hline
Method & \#Param & Guided mesh & IoU (\%) $\uparrow$ & \textit{Lap.} $\downarrow$ & \textit{Cham.} $\downarrow$ \\
\hline 
COAP \cite{bib:COAP} & 138K & MANO & 94.08 & 2.371 & 8.564 \\
COAP & 138K & MANO-HD & 94.01 & 2.348 & 7.694  \\
COAP$^*$ & 287K & MANO-HD & 95.08 & 2.339 & 7.405  \\
\hline
PairOF $^\dagger$ & 256K & MANO-HD & 96.06 & 2.288 & \bf 6.998 \\
PairOF (ours) & 237K  & MANO-HD& \bf 96.32 & \bf 2.281 & 7.151  \\
\hline
\end{tabular}
\caption{Effects of MANO-HD and PairOF. $^*$: w/ wider-MLP decoder; $^\dagger$: w/ Transformer-based decoder.}
\label{tab:geo}
\end{table}

\begin{figure}[t]
\begin{center}
\includegraphics[width=\linewidth]{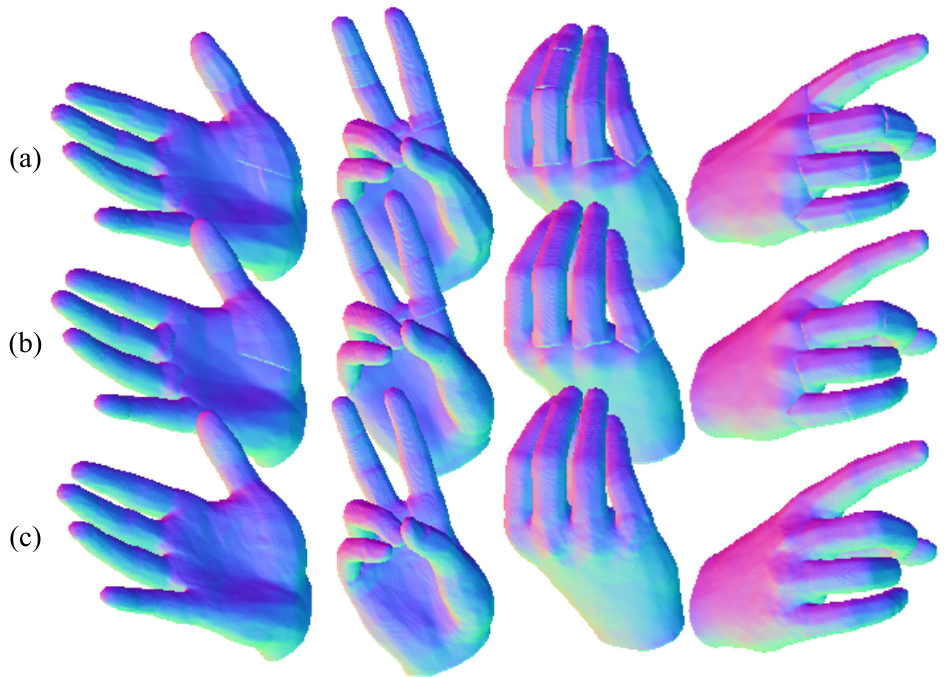}
\caption{Effects of MANO-HD and PariOF. (a) COAP w/ MANO. (b) COAP w/ MANO-HD. (c) PariOF w/ MANO-HD.}
\label{fig:occ}
\end{center}
\vspace{-0.5cm}
\end{figure}

\begin{figure}[t]
\begin{center}
\includegraphics[width=\linewidth]{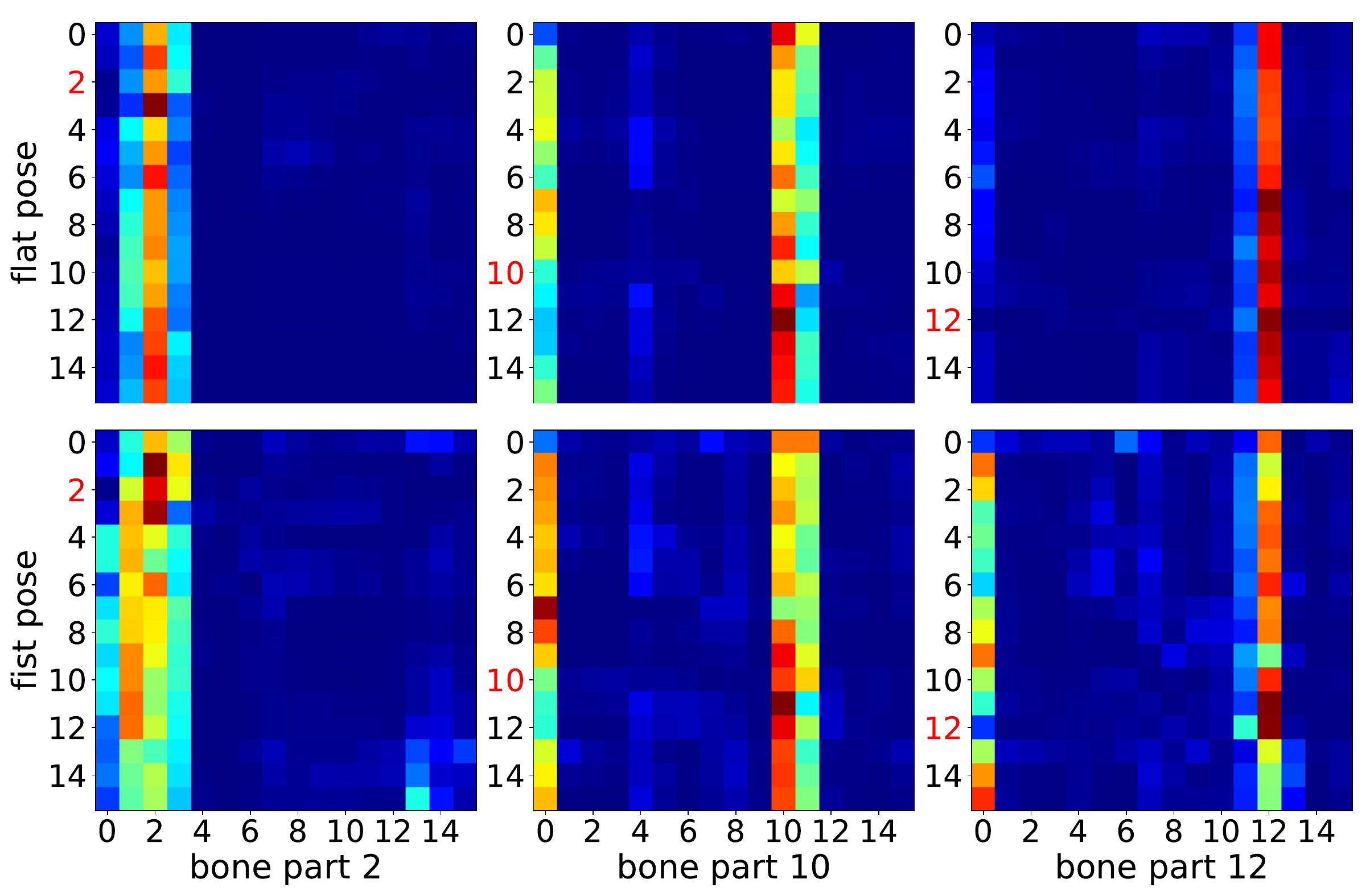}
\caption{Attention map for fusing part-wise geometry encodings. The rows with red indices contribute to global occupancy value.}
\label{fig:att}
\end{center}
\vspace{-0.3cm}
\end{figure}

\begin{figure*}[t]
\begin{center}
\includegraphics[width=\linewidth]{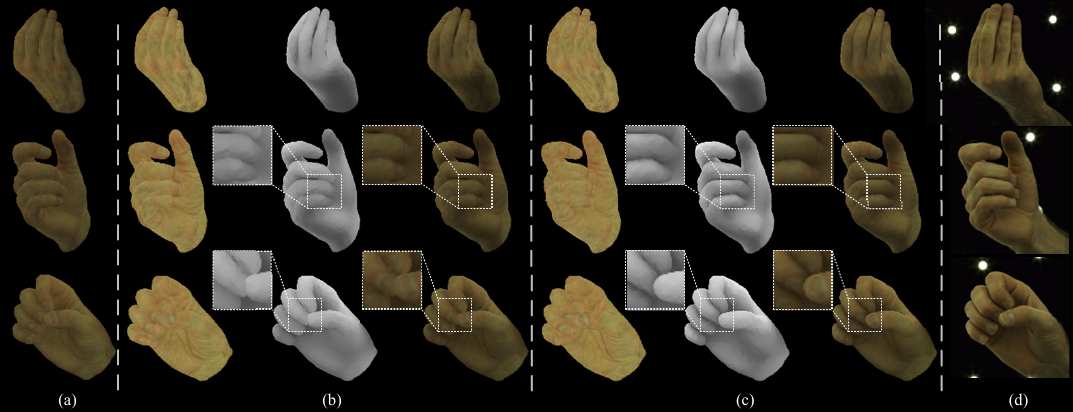}
\caption{Effects of the disentangled albedo and illumination fields in SelF. (a) Coupled albedo and illumination. (b,c) Disentangled albedo and illumination; directed soft occupancy is not involved in (b); from left to right: albedo, illumination, shaded image. (d) Ground truth.}
\label{fig:shadow}
\end{center}
\vspace{-0.3cm}
\end{figure*}

\vspace{-0.3cm}
\paragraph{Metrics}
Following COAP \cite{bib:COAP}, IoU is used to evaluate the occupancy field. We also employ Laplacian smooth (\textit{Lap.}) and Chamfer distance (\textit{Cham.}) to evaluate the mesh quality, the latter of which is formulated as the minimal distance between the vertices extracted from occupancy field \cite{bib:mc} and the guided mesh faces.
\textit{Lap.} and \textit{Cham.} are presented in $10^{-4}$m. Consistent with HumanNeRF \cite{bib:HumanNeRF}, we report LPIPS \cite{bib:lpips}, PSNR, and SSIM \cite{bib:ssim} to reflect image similarity as the metrics of rendering quality. All evaluation data are with novel poses that are unseen in training.

\subsection{Evaluation on Geometry Performance}

\paragraph{Comparison with Prior Arts}
PairOF and COAP \cite{bib:COAP} use the same encoder but different decoders, so their comparison can reveal the effect of our part-pair-wise decoding. The results of COAP are from the officially released code, and we re-train models on InterHand2.6M dataset. Because our local-pair decoder is larger than that of COAP, we enlarge MLP width for comparable model size (denoted as COAP$^*$). Referring to Table~\ref{tab:geo}, an occupancy field guided by MANO-HD has a smoother surface (lower \textit{Lap.}) and higher fidelity (lower \textit{Cham.}), so there are benefits of MANO-HD over MANO in guiding an implicit function. Moreover, PairOF can improve all metrics by a large margin. As shown in Fig.~\ref{fig:occ}, MANO-HD can improve overall smoothness, while PairOF exhaustively eliminates non-smooth part connections to achieve across-part consistency.

\vspace{-0.3cm}
\paragraph{Comparison with Transformer-Based Decoder}
To verify the local-pair prior knowledge in PairOF, the Transformer \cite{bib:Transformer} technique is employed as the decoder, where self-attention can adaptively fuse part-wise geometry without inductive prior. As shown in Table~\ref{tab:geo}, the local-pair decoder performs on par with the Transformer-based decoder. To unveil the effect of self-attention, we delve deep into attention-based feature fusion based on two representative hand poses (\textit{i.e.}, flat and fist poses). At first, we extract mesh vertices from the occupancy field, each of which comes with respective attention maps. Then, we find the part that has the maximal occupancy value for each vertex and gather vertices into groups accordingly. Each group of vertices can reflect the property of a bone part, and we show their average attention maps in Fig.~\ref{fig:att}. Because of the maximum operator in part-wise geometry fusion, only one part contributes to the global occupancy value. That is, for part $b$, we should focus on the $b$th row (red number in Fig.~\ref{fig:att}). Consequently, attention-based fusion is consistent with our local-pair design. For example, the attention map fuses parts $b_0,b_{10},b_{11}$ to evolve the encoding of part $b_{10}$. Therefore, the design concept of the local-pair decoder is evident. Nevertheless, the Transformer-based decoder is not efficient enough because the attention map contains \textit{vertical} patterns instead of \textit{diagonal} ones. That is, meaningless computations are introduced by the Transformer, despite they do not contribute to global occupancy. For example, referring to ``fist pose'' and ``bone part 12'' in Fig.~\ref{fig:att}, the attention map integrates encodings of parts $b_0,b_{12}$ (instead of $b_1$) for the prediction of part $b_1$. The reason behind this is that the \textit{inside} property is exclusively enhanced, and vertices belonging to part $b_{12}$ also have inside properties to part $b_0$ under the fist pose. Refer \textit{suppl. material} for more details and part indices.

\subsection{Evaluation on Rendering Quality}

\paragraph{Ablation Study on Shape Fitting}
We fit personalized hand shape with $\mathcal M_{shape}$. For comparison, $\mathcal M_{shape}$ is replaced with $\boldsymbol{\beta}$-based shape fitting, where $\boldsymbol{\beta}$ is the annotation in InterHand2.6M dataset. As shown in Table~\ref{tab:render_ablation}, $\boldsymbol{\beta}$-based shape induces poor rendering quality, while our method brings a significant improvement. Therefore, the shape fitting capacity of our proposed MANO-HD is confirmed.

\begin{table}[t]
\small
\setlength\tabcolsep{4.5pt}
\centering
\begin{tabular}{c c c | c c c}
\hline
Shape fit. & Illum. & Dir. occ. & LPIPS $\downarrow$ & PSNR $\uparrow$ & SSIM $\uparrow$ \\
\hline
& & & 0.1268 & 26.53 & 0.8692\\
$\checkmark$ & & & 0.1113 & 27.32 & 0.8830 \\
$\checkmark$ & $\checkmark$ & & 0.1063 & 28.02 & 0.8903    \\
$\checkmark$ & $\checkmark$ & $\checkmark$ & \bf 0.1035 & \bf 28.23 & \bf 0.8941 \\
\hline
\end{tabular}
\caption{Effects of shape fitting (Shape fit.), illumination field (Illum.), and directed soft occupancy (Dir. occ.) on rendering quality. Results are from Interhand2.6M \textit{test/Capture0}.}
\label{tab:render_ablation}
\vspace{-0.1cm}
\end{table}

\vspace{-0.3cm}
\paragraph{Ablation Study on SelF}
Referring to Fig.~\ref{fig:shadow}(a) and (b), the disentanglement of albedo and illumination fields can improve the rendering reality by introducing shadow patterns. Moreover, directed soft occupancy can further elevate the illumination representation through ray-based occlusion estimation. Referring to Fig.~\ref{fig:shadow}(c), it is remarkable that the shadow patterns on the palm and fingers are more faithful with fewer artifacts when compared to Fig.~\ref{fig:shadow}(b). In Table~\ref{tab:render_ablation}, the illumination field and directed soft occupancy lead to quantitative improvements in rendering metrics, indicating our SelF is conducive to realistic rendering.

\begin{table*}[t]
\small
\centering
\begin{tabular}{c || c c c || c c c || c c c  }
\hline
\multirow{2}{*}{Method} & \multicolumn{3}{c||}{\textit{test/Capture0}} & \multicolumn{3}{c||}{\textit{test/Capture1}} & \multicolumn{3}{c}{\textit{val/Capture0}} \\ 
 & LPIPS $\downarrow$ & PSNR $\uparrow$ & SSIM $\uparrow$ & LPIPS $\downarrow$ & PSNR $\uparrow$ & SSIM $\uparrow$ & LPIPS $\downarrow$ & PSNR $\uparrow$ & SSIM $\uparrow$ \\
\hline 
SelfRecon \cite{bib:SelfRecon} & 0.1421 & 26.38 & 0.8786 & 0.1389 & 25.18 & 0.8758 & 0.1490 & 25.78 & 0.8687 \\ 
HumanNeRF \cite{bib:HumanNeRF} & 0.1145 & 27.64 & 0.8836 & 0.1177 & 26.31 & 0.8803 & 0.1192 & 27.80 & 0.8816 \\
\hline
ours & \bf 0.1035 & \bf 28.23 & \bf 0.8941 & \bf 0.1076 & \bf 26.56 & \bf 0.8902 & \bf 0.1062 & \bf 28.04 & \bf 0.8900 \\ 
\hline
\end{tabular}
\caption{Rendering quality comparison among our HandAvatar and prior arts on the InterHand2.6M dataset.}
\label{tab:render}
\vspace{-0.1cm}
\end{table*}

\begin{figure}[t]
\begin{center}
\includegraphics[width=\linewidth]{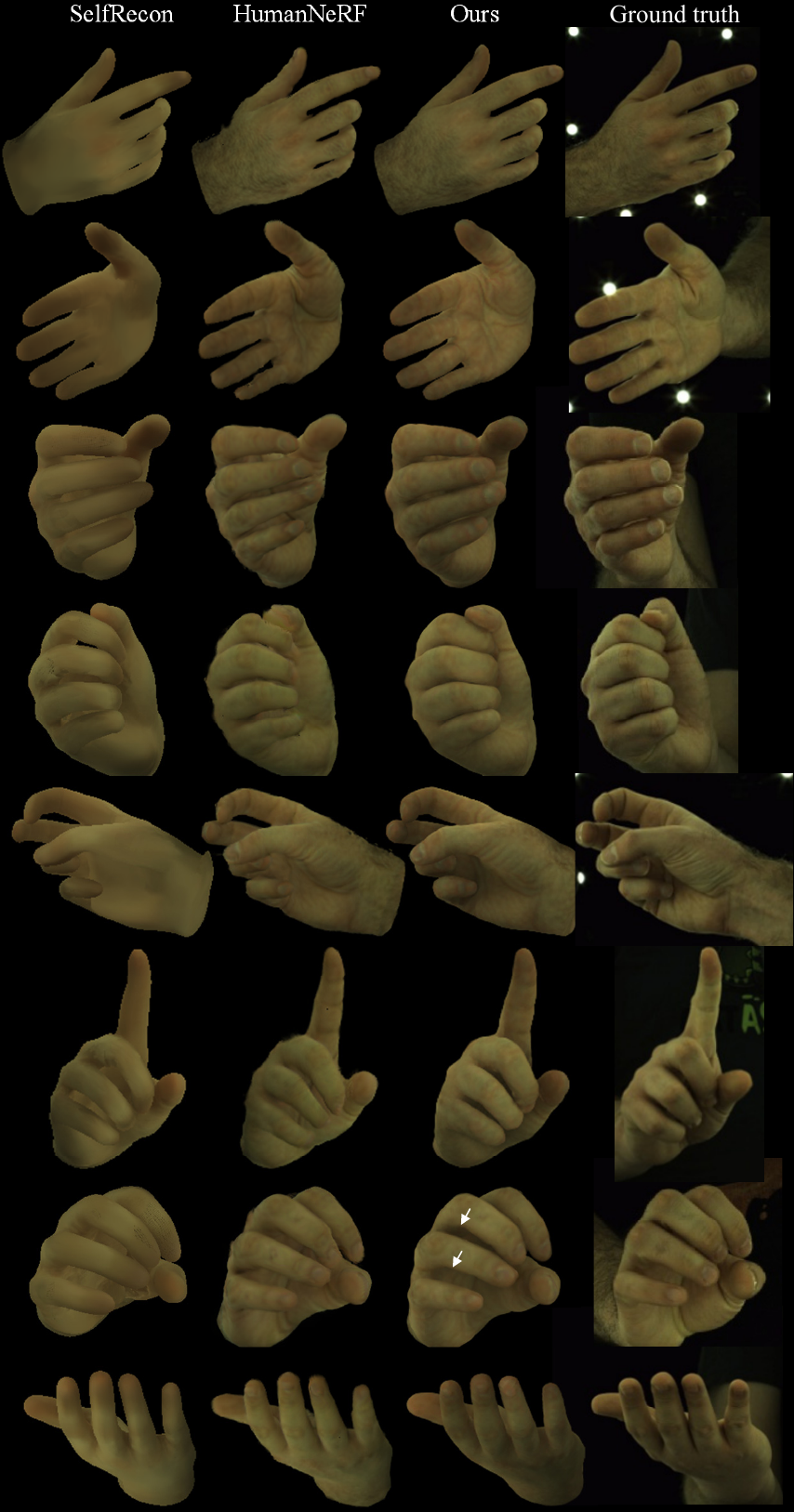}
\caption{Visualization results of SelfRecon \cite{bib:SelfRecon}, HumanNeRF \cite{bib:HumanNeRF}, and our HandAvatar on free-pose animation and rendering.}
\label{fig:comp}
\end{center}
\vspace{-0.5cm}
\end{figure}

\begin{figure}[t]
\begin{center}
\includegraphics[width=\linewidth]{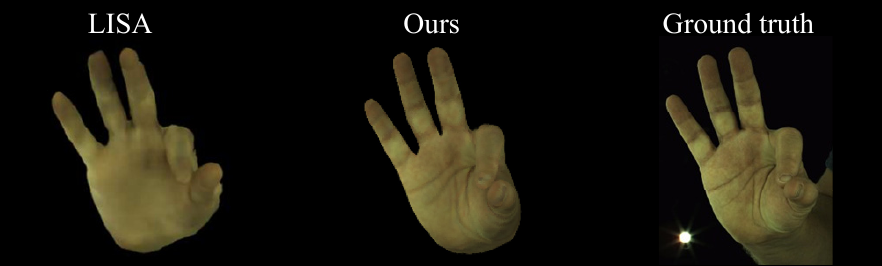}
\caption{Visualization results of LISA \cite{bib:LISA} and our HandAvatar. The result of LISA is cropped from their original paper.}
\label{fig:lisa}
\end{center}
\vspace{-0.5cm}
\end{figure}

\vspace{-0.3cm}
\paragraph{Comparison with Prior Arts}
We compare HandAvatar with previous monocular methods HumanNeRF \cite{bib:HumanNeRF} and  SelfRecon \cite{bib:SelfRecon}, both of which are from officially released codes and re-trained on InterHand2.6M dataset. SelfRecon uses the surface-based rendering \cite{bib:idr} method, and its representation of texture detail is not good enough, as shown in Fig.~\ref{fig:comp}. HumanNeRF and our method leverage the volume rendering method \cite{bib:NeRF}, which can produce realistic hand texture. However, limited by inverse skinning deformation, HumanNeRF cannot cope with self-contact that commonly occurs in hand animation. As shown in Fig.~\ref{fig:comp}, HumanNeRF has corrupted geometry when fingers contact with each other. In contrast, benefiting from our PairOF, HandAvatar has the advantage of free-pose animation while at the same time maintaining geometry fidelity. In addition, both SelfRecon and HumanNeRF employ entangled albedo and illumination for color prediction, so the shadow on hand is hard to be aware of, as shown in the 3rd-8th rows of Fig.~\ref{fig:comp}. On the contrary, our HandAvatar leads to a superiority in modeling illumination and shadow patterns caused by articulated self-occlusion. This is attributed to the disentanglement of albedo and illumination brought by our proposed SelF. Further, referring to the white arrows in Fig.~\ref{fig:comp}, we can present varying shadow intensities for different self-occlusion situations, thanks to our directed soft occupancy.
For quantitative comparison, we train models on three video sequences and achieve the best results in all metrics, as shown in Table~\ref{tab:render}.

As the most related work to this paper, LISA \cite{bib:LISA} is trained on non-released multi-view data, and its models/codes remain unavailable. Thereby, we compare LISA based on the result reported in their original paper. As shown in Fig.~\ref{fig:lisa}, LISA has difficulty in capturing accurate hand pose with a learnable skinning-based deformation. Besides the faithful shape and pose reconstruction, our rendered texture details are more realistic than that of LISA.

In addition to rendering fidelity, HandAvatar also provides a route for appearance editing as shown in Fig.~\ref{fig:fig1}.
\section{Conclusions}
In this work, we present a novel hand representation called HandAvatar for free-pose animation and rendering. First, we extend MANO to MANO-HD as a high-resolution topology structure to improve the shape-fitting capacity of hand mesh. Subsequently, PairOF with a local-pair decoder is developed, which can generate an across-part consistent occupancy field. Furthermore, we propose SelF, the first approach to model hand texture under articulated self-occlusion, to disentangle hand albedo and illumination. Extensive experiments demonstrate our superior results on free-pose hands animation and rendering. We believe our method paves a new way for dynamic hand representation.

\vspace{-0.4cm}
\paragraph{Limitation and future works}
For affordable computational costs, we use the directed soft occupancy on view direction to estimate the irradiance. This could lead to view-direction-dependent shadow patterns. Thus, the improved illumination field is worthy of ongoing exploration.

{\small

}

\clearpage
\appendix

\section{MANO-HD}

\paragraph{MANO.}
MANO can be driven with parameters $\boldsymbol{\beta}\in\mathbb R^{10}$ and $\boldsymbol{\theta}\in\mathbb R^{B\times 3}$ ($B=16$ indicates the number of per-bone parts), where $\boldsymbol{\beta}$ is the coefficients of a shape PCA bases while $\boldsymbol{\theta}$ represents joint rotations in axis-angle form. Mean template mesh is deformed to match different shapes:
\begin{equation}
\label{eq:shape}
\begin{array}{l}
    \tilde{\mathbf V} =\bar{\mathbf V}+\mathcal B_s(\boldsymbol\beta) + \mathcal B_p(\boldsymbol\theta) \\[5pt]
    \mathbf J =\mathcal J(\bar{\mathbf V}+\mathcal B_s(\boldsymbol\beta)),
\end{array}
\vspace{-0.1cm}
\end{equation}
where $\bar{\mathbf V},\mathcal B_s,\mathcal B_p$ are template vertices and shape/pose blendshapes. Canonical joint locations $\mathbf J\in\mathbb R^{B\times 3}$ are given with the regressor $\mathcal J$.

Then, bone transformation matrix $\mathbf G_b\in\mathbb R^{4\times 4}$ is computed along the kinematic chain $\mathcal K$ with the Rodriguez formula $\mathcal R$:
\begin{equation}
\label{eq:trans}
\begin{array}{l}
    \mathbf G_b(\boldsymbol\theta, \mathbf{J})=\prod_{j \in \mathcal K(b)}\left[\begin{array}{c|c}
    \mathcal R(\boldsymbol\theta_j) & \mathbf{J}_j \\
    \hline \mathbf{0} & 1
    \end{array}\right]
\end{array}
\vspace{-0.1cm}
\end{equation}

Finally, linear blend skinning is used to pose vertices with skinning weights $\mathbf W\in\mathbb R^{V\times B}$ ($V$ denotes the number of vertices) as follows,
\begin{equation}
\label{eq:skin}
\begin{array}{l}
\mathbf V_i=\sum_{b=1}^B \mathbf{W}_{b, i} \mathbf G_b(\boldsymbol\theta, \mathbf{J}) G_k(\mathbf{0}, \mathbf{J})^{-1} \tilde{\mathbf V}_i.
\end{array}
\vspace{-0.1cm}
\end{equation}

\paragraph{Optimization of MANO-HD.}
Following \cite{bib:NHA}, we subdivide the MANO template mesh to obtain a high-resolution version with 12,337 vertices and 24,608 faces. Nevertheless, articulated dynamic mesh subdivision is a non-trivial task, and mesh skinning operation is likely to introduce artifacts to deformed mesh. Thus, we optimize upsampled skinning weights $\mathbf W^{HD}\in\mathbb R^{V^{HD}\times B}$ to eliminate dynamic artifacts under various hand poses using energy functions as follows,
\begin{equation}
\begin{array}{l}
    \mathcal L_{l_0} = \sum_{i=1}^{B\cdot V^{HD}}(1-e^{-\eta\mathbf W^{HD}_i}) \\[5pt]
    \mathcal L_{lap} = \frac{1}{V^{HD}}\sum_{i=1}^{V^{HD}}\sum_{j\in\mathbb N(i)}\frac{1}{\omega}\mathbf ||\mathbf V^{HD}_i-\mathbf V^{HD}_j||_2 \\[5pt]
    \mathcal L_{surf} = \mathrm{Cham}(\mathbf V^{HD}, \mathbf F),
\end{array}
\label{eqn:manohd}
\end{equation}
where $\mathcal L_{l_0}$ is approximated $l_0$ norm constraint \cite{bib:l0} to produce sparse skinning weights. $\mathcal L_{lap}$ is the Laplacian term for mesh smoothness, where $\mathbb N(\cdot)$ represents vertex neighborhood and $\omega$ is the normalization factor. The function $\mathrm{Cham(\cdot,\cdot)}$ computes the chamfer distance between MANO-HD mesh vertices $\mathbf V^{HD}$ and the MANO mesh faces $\mathbf F$. The overall energy function is given as $\mathcal L_{HD}=\lambda_{l_0}\mathcal L_{l_0}+\lambda_{lap}\mathcal L_{lap}+\lambda_{surf}\mathcal L_{surf}$ with balance term $\lambda$.

\vspace{-0.3cm}
\paragraph{Implement Details of MANO-HD}
For $\mathbf W^{HD}$ optimization, we adopt the original MANO dataset \cite{bib:MANO} with 1,554 pose parameters for training and evaluation. We randomly compose and interpolate finger-level rotations for data augmentation. The training process lasts 3,000 steps with a batch size of 1,024. The learning rate begins at $10^{-5}$ and decreases with exponential decay. We use $\mathcal L_{HD}$ as the objective and adjust hyperparameters to balance multiple energy terms as $\eta=100,\lambda_{l_0}=0.01,\lambda_{lap}=1,\lambda_{surf}=10$.

\begin{table}[t]
\small
\renewcommand{\arraystretch}{1.1}
\centering
\begin{tabular}{c | c c c}
\hline
Method  & \textit{Lap.} & \textit{Cham.} & $l_0$ norm (\%) \\
\hline 
MANO & 23.31 & - & 16.29 \\
MANO-HD w/o $\mathbf W^{HD}$ opt. & 1.923 & 7.014 & 17.97 \\ 
MANO-HD w/o $\mathcal L_{l_0}$ & 1.576 & 7.170 & 44.15 \\ 
MANO-HD   & 1.753 & 7.039 & 16.89 \\ 
\hline
\end{tabular}
\caption{Effects of the $\mathbf W^{HD}$ optimization (opt.).}
\label{tab:mano}
\end{table}

\begin{figure}[t]
\begin{center}
\includegraphics[width=\linewidth]{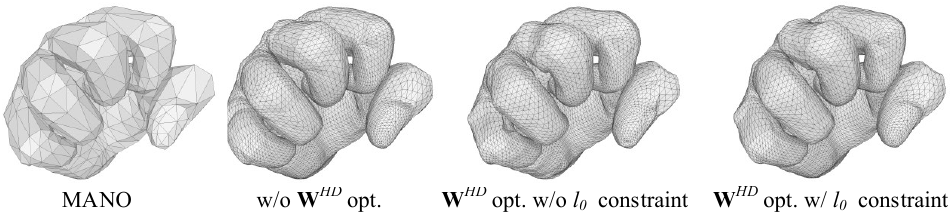}
\caption{Hand mesh comparison under a large deformation.}
\label{fig:manohd}
\end{center}
\vspace{-0.2cm}
\end{figure}

\vspace{-0.3cm}
\paragraph{Effects of MANO-HD}
We use Laplacian smoothing \textit{Lap.} and chamfer distance \textit{Cham.} to reflect the smoothness and accuracy of MANO-HD, whose definition is the same as $\mathcal L_{lap}$ and $\mathcal L_{surf}$ in Eq.~\ref{eqn:manohd}. Both \textit{Lap.} and \textit{Cham.} are presented in $10^{-4}$m. As the MANO surface is not smooth enough, \textit{Lap.} and \textit{Cham.} cannot be simultaneously improved. Besides, we introduce $l_0$ norm as the metric, which is defined as the proportion of non-zero elements in the skinning weights.

Referring to Tab.~\ref{tab:mano}, mesh subdivision can improve \textit{Lap.}, yet incurs artifacts during skinning as shown in the part-connection regions in Fig.~\ref{fig:manohd}. Moreover, $\mathbf W^{HD}$ optimization without $\mathcal L_0$ cannot counteract the issue despite inducing lower \textit{Lap.} value. The reason behind this is the poor sparsity of $\mathbf W^{HD}$. After $\mathcal L_{l_0}$-based optimization, the $l_0$ norm of $\mathbf W^{HD}$ is on par with that of MANO, and the skinning performance is improved (see Fig.~\ref{fig:manohd}).

\section{Network Structures}

\begin{figure}[t]
\begin{center}
\includegraphics[width=0.7\linewidth]{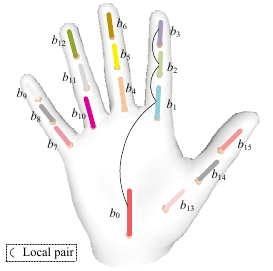}
\caption{Kinematic tree and part indices of the hand. We also show local pairs along the kinematic chain of forefinger.}
\label{fig:kinematics}
\end{center}
\end{figure}

\begin{figure}[t]
\begin{center}
\includegraphics[width=0.8\linewidth]{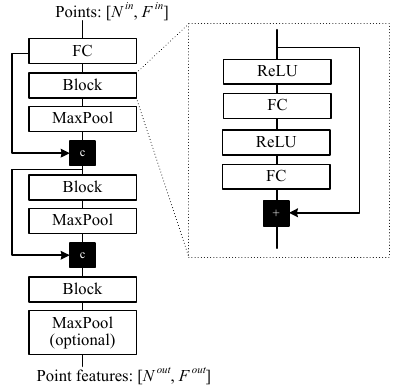}
\caption{PointNet structure. ``c,+'' indicate concatenation and element-wise sum; ``FC'' denotes fully connected layer.}
\label{fig:pointnet}
\end{center}
\vspace{-0.3cm}
\end{figure}

\begin{table}[t]
\small
\renewcommand{\arraystretch}{1.1}
\centering
\begin{tabular}{c | c c c c }
\hline
Name  & Depth & Width & Input size & Output size \\
\hline 
$\mathcal M_{shape}$ & 4 & 128 & 84 & 3 \\
MLP in $\mathcal Q_{pair}$ & 4 & 128 & 64 & 1 \\
$\mathcal M_{albedo}$ & 4 & 256 & 128 & 3 \\
$\mathcal M_{illum}$ & 4 & 256 & 85 & 1 \\
\hline
\end{tabular}
\caption{The details of MLPs.}
\label{tab:mlp}
\end{table}

\paragraph{Kinematic Tree and Local Pair.}
Following the definition of MANO \cite{bib:MANO}, Fig.~\ref{fig:kinematics} shows the bone indices and connections for the hand. In addition, we demonstrate our defined local pairs.

\vspace{-0.3cm}
\paragraph{PointNet Structure.}
The PointNet used in part-space encoder $\mathcal{Q}_{part}$ and local pair decoder $\mathcal{Q}_{pair}$ is shown in Fig~\ref{fig:pointnet}, where $N,F$ denote point amount and feature size. For $\mathcal{Q}_{part}$, $N^{in}=256,F^{in}=6,N^{out}=1,F^{out}=64$. For $\mathcal{Q}_{pair}$, we get rid of the last $\mathrm{MaxPool}$, leading to  $N^{in}=2,F^{in}=64,N^{out}=2,F^{out}=64$. 

\vspace{-0.3cm}
\paragraph{MLP Details.}
Referring to Table~\ref{tab:mlp}, we list MLP details used in this paper.

\vspace{-0.3cm}
\paragraph{Local-Pair Decoder.}

\begin{figure}[t]
\begin{center}
\includegraphics[width=\linewidth]{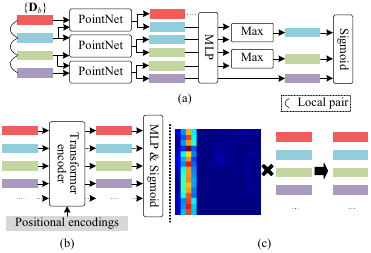}
\caption{(a) Our proposed local-pair decoder. Multiple PointNets/MLPs share weights. (b) Transformer-based decoder for comparison. (c) Attention-based fusion of part-level encodings in Transformer for bone part $b_{2}$. Consistent with Fig.\ref{fig:kinematics}, colors distinguish part-level geometry encodings. For visual conciseness, we only show the process of forefinger.}
\label{fig:decoder}
\end{center}
\vspace{-0.2cm}
\end{figure}

\begin{figure*}[t]
\begin{center}
\includegraphics[width=\linewidth]{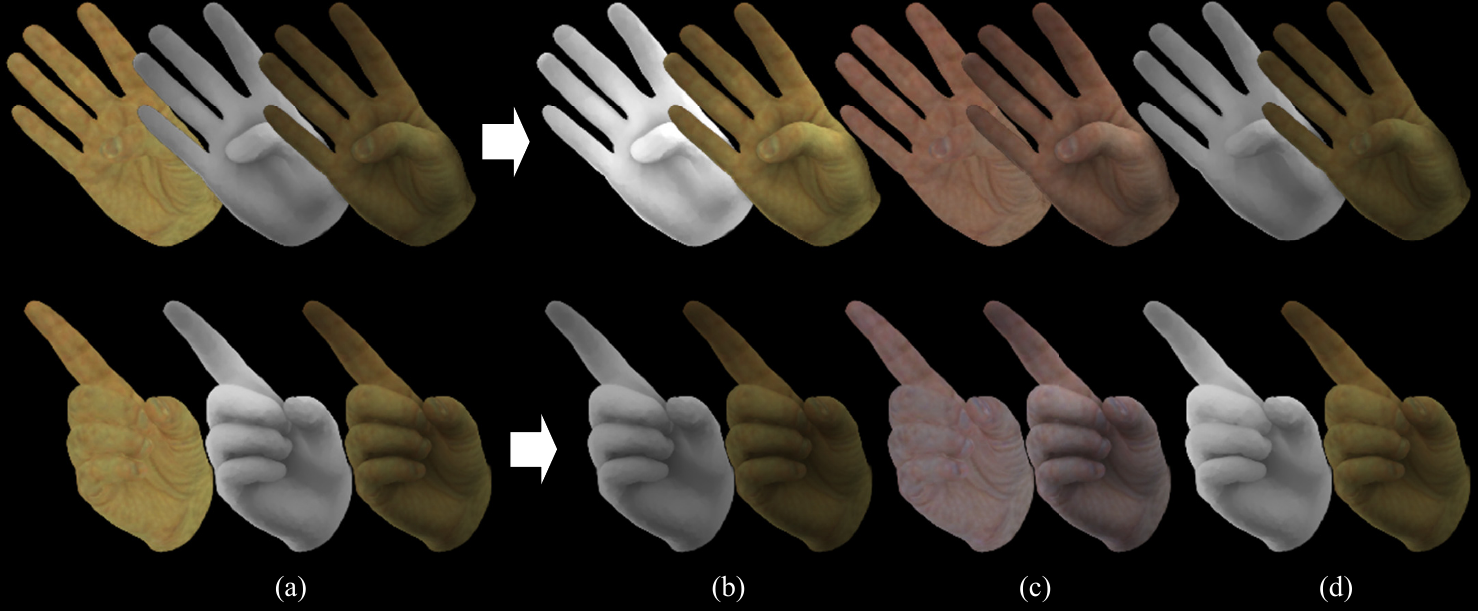}
\caption{Three groups of texture editing. (a) Reconstruction results. From left to right: albedo, illumination, and shaded image. (b) Lighting editing. (c) Albedo editing. (d) Shadow editing.}
\label{fig:edit}
\end{center}
\vspace{-0.3cm}
\end{figure*}

The local pair decoder $\mathcal Q_{pair}$ contains a PointNet and an MLP, whose detailed structures are introduced before. Furthermore, we illustrate how the $\mathcal Q_{pair}$ processes the geometry encodings along the kinematic chain of forefinger, as shown in Fig.~\ref{fig:decoder}(a). First, locally paired encodings are treated as two \textit{points} and fused by the PointNet. Without across-point maxpooling, the shape of PointNet output remains the same as its input. Then, the MLP maps PointNet outputs to the occupancy domain. As a result, a bone part could have multiple occupancy values from multiple local-pair predictions, which are fused by a maximum operator. Because of the feature fusion by the PointNet, part boundaries are blurred and extend to the connection direction. Hence, the maximum operator is used to produce a union of boundary-extended part geometries. Finally, the sigmoid function is employed for occupancy normalization.

\vspace{-0.3cm}
\paragraph{Transformer-Based Decoder.}
To validate the local-pair design, we develop a learning-based method with Transformer. As shown in Fig.~\ref{fig:decoder}(b), all part-level encodings are fed to the Transformer encoder without any inductive prior. Through 4 self-attention blocks, the Transformer can perform adaptive feature fusion. Referring to Fig.~\ref{fig:decoder}(c), the attention map determines the way to select important features, where the effect is consistent with our local-pair decoder. That is, bone parts $b_1,b_2,b_3$ are connected in the attention map for evolving the encoding of part $b_2$.

\section{Training Details}

\paragraph{Video Data.}

\begin{table}[t]
\small
\renewcommand{\arraystretch}{1.1}
\centering
\begin{tabular}{c | c c  }
\hline
Capture   & Training set &  Validation set \\
\hline 
\textit{test/Capture0} & 11,757 & 194 \\
\textit{test/Capture1} & 18,474 & 232 \\ 
\textit{val/Capture0} & 18,340 & 184 \\ 
\hline
\end{tabular}
\caption{Data amount for training and validation sets.}
\label{tab:data}
\vspace{-0.1cm}
\end{table}

For quantitative evaluation, we select three sequences from the InterHand2.6M dataset \cite{bib:InterHand}. Data amount is shown in Table~\ref{tab:data}, where training data are from the \textit{ROM04\_RT\_Occlusion} sequence and validation data are from the \textit{ROM03\_RT\_No\_Occlusion} sequence. Because video frames are highly redundant, validation data are selected by fixed skip steps, and we adjust the steps to assure various hand poses and global rotations can be covered. 

For each frame, we crop the hand region with annotated detection boxes as the ground truth. First, the box is regulated as a square box with 1.3 times expansion. Then, the hand region is cropped and resized to $256\times 256$ resolution.

\vspace{-0.3cm}
\paragraph{Training Settings.}
For PairOF pre-training, the learning rate begins at $5\times 10^{-4}$ and decreases with exponential decay. The training process has 270K steps with a batch size of 32.

For end-to-end training, the learning rate begins at $5\times 10^{-4}$ and decreases with exponential decay. Due to the pre-training, the learning rate of $\mathcal Q_{pair}$ is set to be 10 times smaller. Following \cite{bib:HumanNeRF}, we use a patch strategy for training with a patch size of $32\times 32$. The training process has 50K steps with a batch size of 16.

\section{Texture Editing}

Referring to Fig~\ref{fig:edit}, our HandAvatar supports hand texture editing. Firstly, we change the illumination field by a 1.5- or 0.8-times multiplication, as shown in Fig~\ref{fig:edit}(b). Then, we shift the mean RGB value of the albedo field, leading to the results of Fig~\ref{fig:edit}(c). Besides, we edit shadow in Fig~\ref{fig:edit}(d).  In the top row, the texture is induced by letting thumb-related directed soft occupancy values equal 0. As a result, the self-occluded shadow patterns on the palm become weakened. In the bottom row, we remove the shadow patterns between the middle and ring fingers by setting the directed soft occupancy values as 0 for bone parts $b_5$ and $b_{11}$. The shadow editing results also validate the effect of directed soft occupancy.


\section{Discussion}

\paragraph{The Disentanglement of Albedo and Illumination.}
The disentanglement of albedo and illumination does not require extra regularization. Albedo is known to be independent of hand pose, while illumination depends on hand pose. In SelF, the input of the albedo field is unrelated to the hand pose, while pose-relevant elements are fed into the illumination field. Therefore, with various hand poses as training data, the optimization process would ensure that the illumination is free from the albedo field. In addition, the illumination field outputs a scalar, which cannot model RGB-based (3-channel) albedo.

\begin{figure}[t]
\centering
\includegraphics[width=0.7\linewidth]{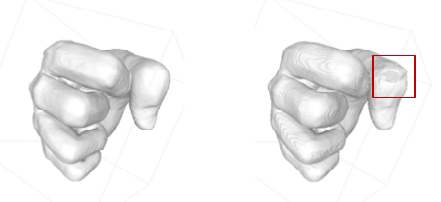}
\caption{PairOF (left) \vs Transformer-based method (right)}
\label{fig:trans}
\vspace{-5pt}
\end{figure}

\paragraph{PairOF \vs Transformer-Based Method}
The motivation of PairOF is to fuse part-level geometry encodings to eliminate the shape inconsistency in the area of part connections. Despite similar numerical results in Tab.~2 of the  main text, our PairOF is more effective in feature fusion than the Transformer-based method. That is, hand bone connections are unchangeable, and PairOF can use this prior knowledge, leading to a more effective feature fusion than self-attention. As a result, self-attention cannot visually achieve our motivation as shown in Fig.~\ref{fig:trans}. 
Also, we are more efficient in terms of learning, \ie, the convergence of the Transformer-based method is slower and more data-hungry. 

\section{Limitations and Future Works}
Besides the limitation discussed in the main text, HandAvatar can be further improved from the following perspectives in the future. First, full lighting editing is worthy of future research. For example, despite the lighting editing as shown in Fig~\ref{fig:edit}(b), it is hard to edit or add a point light in the illumination field. Second, the representation of specular effects in hand appearance is another interesting topic. To achieve this goal, hand surface properties with BRDF should be explored. Third, the hand geometry demonstrated in Fig. 5 in the main text is the result of PairOF pre-training. After end-to-end training with texture losses, the PairOF could produce a non-smooth surface with geometry wrinkles. This is caused by the hand-pose annotation error of InterHand2.6M. As a result, the PairOF could produce fragile hand geometry to compensate for this error.

\end{document}